\documentclass[times,twocolumn,final,authoryear]{elsarticle}

\usepackage{medima_Arxiv}
\journal{Accepted for publication at Medical Image Analysis}

\usepackage{framed,multirow}

\usepackage{amssymb}
\usepackage{latexsym}

\usepackage{url}
\usepackage{xcolor}

\usepackage{hyperref}
\usepackage[switch]{lineno}
\usepackage{wrapfig}
\usepackage{booktabs}
\usepackage{soul}
\usepackage{verbatim}
\usepackage[cmex10]{amsmath}
\definecolor{newcolor}{rgb}{.8,.349,.1}
\usepackage{relsize}
\def \Em{{\mathbb{E}}}
\def \Rm{{\mathbb{R}}}

\def \Im{{\mathbb{I}}}
\def \abf{{\mathbf a}}
\def \Abf{{\mathbf A}}

\def \hbf{{\mathbf h}}

\def \sbf{{\mathbf s}}
\def \Sbf{{\mathbf S}}

\def \xbf{{\mathbf x}}
\def \Xbf{{\mathbf X}}

\def \0bf{{\mathbf 0}}

\def \alphabf{{\boldsymbol{\mathlarger{\mathlarger{\alpha}}}}}

\def \Ncal{{\mathcal N}}

\def \Fcal{{\mathcal F}}
\def \Ecal{{\mathcal E}}

\def \Qcal{{\mathcal Q}}
\def \Gcal{{\mathcal G}}
\def \Lcal{{\mathcal L}}
\def \Vcal{{\mathcal V}}

\begin{document}

\verso{R. Selvan \textit{et~al.}}

\begin{frontmatter}

\title{Graph Refinement based Airway Extraction using Mean-Field Networks and Graph Neural Networks}

\author[1]{Raghavendra {Selvan}\corref{cor1}}
\cortext[cor1]{Corresponding author at Universitetsparken 1, Denmark-2100. raghav@di.ku.dk \\
		Source code for the two graph reginement models will be made available here \url{https://github.com/raghavian/graph_refinement}
}
\author[uva]{Thomas {Kipf}} 
\author[uva,cifar]{Max Welling}
\author[emc]{Antonio Garcia-Uceda Juarez} 
\author[rig]{Jesper H Pedersen} 
\author[ku]{Jens Petersen} 
\author[ku,emc]{Marleen de Bruijne}

\address[ku]{Department of Computer Science, University of Copenhagen, Denmark}
\address[uva]{Informatics Institute, University of Amsterdam, The Netherlands}
\address[cifar]{Canadian Institute of Advanced Research}
\address[emc]{Biomedical Imaging Group Rotterdam, Department of Radiology and Nuclear Medicine, Erasmus MC, Rotterdam, The Netherlands}
\address[rig]{Department of Thoracic Surgery, Rigshospitalet, University of Copenhagen, Denmark}

\received{\today}

\begin{abstract}
Graph refinement, or the task of obtaining subgraphs of interest from over-complete graphs, can have many varied applications. In this work, we extract trees or collection of sub-trees from image data by, first deriving a graph-based representation of the volumetric data and then, posing the tree extraction as a graph refinement task. We present two methods to perform graph refinement. First, we use mean-field approximation (MFA) to approximate the posterior density over the subgraphs from which the optimal subgraph of interest can be estimated. Mean field networks (MFNs) are used for inference based on the interpretation that iterations of MFA can be seen as feed-forward operations in a neural network. This allows us to learn the model parameters using gradient descent. Second, we present a supervised learning approach using graph neural networks (GNNs) which can be seen as generalisations of MFNs. Subgraphs are obtained by training a GNN-based graph refinement model to directly predict edge probabilities. We discuss connections between the two classes of methods and compare them for the task of extracting airways from 3D, low-dose, chest CT data. We show that both the MFN and GNN models show significant improvement when compared to one baseline method, that is similar to a top performing method in the EXACT'09 Challenge, and a 3D U-Net based airway segmentation model, in detecting more branches with fewer false positives.

\end{abstract}



\end{frontmatter}



\section{Introduction}
\label{sec:intro}

Tree structures occur naturally in many places and play vital anatomical roles in the human body. Segmenting them in medical images can be of immense clinical value. Airways, vessels, and neurons are some such structures that have been studied extensively from a segmentation point of view~\citet{lesage2009review,donohue2011automated,pu2012ct,lo2012extraction}. {The tree nature of these structures can be useful in clinical studies. For instance, performing coronary catheterisation procedures like percutaneous coronary intervention~\citet{serruys2009percutaneous} and planning bronchoscopy~\citet{kiraly2004three} rely on high quality 3D segmentation of coronary and airway trees, respectively. Further, labeling of airway generations is widely used to analyse airway branches at specific generations and compare them across different subjects~\citet{feragen2012hierarchical,petersen2011longitudinal} in relation diseases like cystic fibrosis~\citet{wielputz2013automatic} and chronic obstructive pulmonary disease (COPD))~\citet{smith2018human}; analyses like these are dependent on obtaining high quality segmentations.  
}

Many widely used methods for vascular and airway tree segmentation tasks are sequential in nature i.e, they start from one location (a seed point) and segment by making successive local decisions~\citet{lesage2009review,pu2012ct,lo2012extraction}. For instance, in the EXACT'09 airway segmentation challenge~\citet{lo2012extraction}, 10 out of the 15 competing methods used some form of region growing to make the sequential segmentation decisions and the remainder of the methods were also sequential. The methods in~\citet{lo2009airway, selvan2018extracting} are sequential but do not rely on making local decisions; these methods utilise more global information in making the segmentation decisions. Methods that rely primarily on sequential and/or local  decisions are susceptible to local anomalies in the data due to noise and/or occlusion and can possibly miss many branches or entire sub-trees.

Graph-based methods have previously been used for the extraction of vessels~\citet{orlando2014learning}, airways~\citet{graham2010robust, bauer2015graph} and neurons~\citet{turetken2016reconstructing}, predominantly in a non-sequential setting. In~\citet{orlando2014learning}, a pixel-level conditional random field (CRF) based model is presented, with parameterised node and pairwise potentials over the local neighbourhoods to segment 2D retinal blood vessels. The parameters of this CRF model are learned from the training data using support vector machines. Scaling these pixel-level CRF models to 3D data and performing inference using them can be expensive; instead, using nodes with higher level information so as to sparsify the data can be an efficient approximation strategy. In~\citet{bauer2015graph}, a tube detection filter is used to obtain candidate airway branches. These candidate branches are represented as nodes in a graph and airway trees are reconstructed using a two-step graph-based optimisation procedure. This sequential two-step optimisation introduces the possibility of missing branches or sub-trees, and a global optimisation procedure is desirable. In~\citet{turetken2016reconstructing}, the image data is first processed to obtain local regions of interest using a tubularity measure and maximum tubularity points are selected as nodes of the graph. Nodes within a certain neighbourhood are linked using minimal paths to obtain the graph edges. Several expressive feature descriptors for segments of these edges are computed and used as input to a path classifier which further assigns weights to these edges. Finally, subgraphs of interest are recovered using integer linear programming. The emphasis of the method in~\citet{turetken2016reconstructing} is to obtain complex node and edge features, and then use of a global optimisation to reconstruct structures of interest. 


In this work, we also take up a graph-based approach, to overcome some of the shortcomings of the aforementioned methods, by formulating extraction of tree-like structures from volumetric image data as a graph refinement task. The input image data is first processed to obtain a graph-like representation, comprising nodes with information extracted from the local image neighbourhoods. This graph based representation of image data reduces the computational expense, in contrast to the pixel-level CRFs used in~\citet{orlando2014learning}, while also abstracting the tree segmentation task to a higher level than pixel classification. The preprocessed input graphs are initially over-connected based on simple neighbourhood criteria and then the connectivity is refined to obtain the optimal subgraphs that correspond to the structures of interest. When compared to the method in~\citet{bauer2015graph}, the proposed model uses a single global optimisation procedure which removes the chance of missing branches or sub-trees in the intermediate optimisation step. And, when compared to the method in~\citet{turetken2016reconstructing}, we  utilise relatively simpler node features, unweighted edges and extract the subgraphs of interest based on the global connectivity. We propose two approaches to solve graph refinement task in these settings, using : 1) Mean-Field Networks (MFNs) 2) Graph Neural Networks (GNNs).  

In the first proposed method, graph refinement is posed as an approximate Bayesian inference task solved using mean-field approximation (MFA)~\citet{jaakkola1998improving, wainwright2008graphical}. The posterior density over different subgraphs is approximated with a simpler distribution and the inference is carried out using MFA. We introduce parameterised node and pairwise potentials that capture behaviour of the optimal subgraph corresponding to the underlying tree structure and obtain MFA update equations within the variational inference framework~\citet{beal2003variational}. By unrolling the MFA update iterations as layers in a feedforward network, we demonstrate the use of gradient descent to learn parameters of this model and point it out to be Mean-Field Network as was initially suggested in~\citet{li2014mean}. We extend the previously published conference work in~\citet{selvan2018mean} in this paper by performing more comprehensive experiments and presenting a thorough comparison with GNNs.

In the second proposed method, graph refinement is performed using Graph Neural Networks. GNNs are a class of recurrent neural networks operating directly on graph-structured data~\citet{scarselli2009graph,li2015gated} and are now seen as an important step in generalising deep learning models to non-Euclidean domains~\citet{bronstein2017geometric}. 
{Several closely related formulations of GNNs are prevalent in the literature that treat them as generalisations of message passing algorithms~\citet{gilmer2017neural,kipf2018neural} or as attempts at generalising convolutional neural networks to the domain of graphs~\citet{bronstein2017geometric}. However, the down-sampling and up-sampling operations that work well with convolutional neural networks like in~\citet{ronneberger2015u} are not straight-forward to implement on graphs; recent works such as in~\citet{ying2018hierarchical} introduce a differentiable pooling operator that can be used to obtain down-sampled versions of graphs to allow GNNs to learn at multiple resolutions of the graphs. Another view of GNNs is to treat them as deep learning models that can learn useful representations of nodes, edges or even of sub-graphs and graphs~\citet{kipf2016variational,hamilton2017inductive}. More recently, GNNs have also been used to model and learn inter-dependencies in data for relational reasoning~\citet{battaglia2018relational}. Surveys such as in~\citet{hamilton2017representation,zhou2018graph} capture several of the accelerated advancements in GNNs; while the former takes up a representation learning approach discerning different models based on the embeddings learnt, the latter provides a technical summary of the different prevalent GNNs. }

{In this work, we utilise the GNN model first introduced in~\citet{kipf2016variational}. Our proposed GNN model is formulated as a link/edge prediction model, which takes an over-complete graph as input and predicts the refined sub-graph that corresponds to the structure of interest in a supervised setting. The model used has similarities to other GNN-based models that have recently been shown to be conducive to model and learn interactions between nodes and have seen successful applications in modeling interactions of objects in physical environments~\citet{battaglia2016interaction,kipf2018neural}.}
In the presented work, the graph refinement task itself is solved in a supervised setting by jointly training a GNN-based encoder that learns edge embeddings based on the over-complete input graph. At the final layer, we use a single layer perceptron based decoder to obtain the probability of edge connections. The idea of using GNNs for graph refinement was initially proposed in our earlier work in~\citet{selvan2018extraction}, where a GNN-based encoder was used to learn node embeddings and a pairwise decoder was used to predict the edge probabilities. Using node embeddings to predict edge probabilities proved to be inadequate, which we have now addressed in this work by predicting edge probabilities from learnt edge embeddings instead.

In addition to proposing MFNs and GNNs as two methods to solve the graph refinement tasks, we also study connections between them. In the case of MFN model, the node and pairwise potentials are hand-crafted, incorporating useful prior knowledge. With only a handful of parameters the MFN model requires little supervision and can be seen as an intermediate between a model-based solution and the fully end-to-end training model based on GNNs. On the other hand, the GNN models can be seen as generalisation of message passing algorithms used for inference in probabilistic graphical models~\citet{wainwright2008graphical} such as MFNs. When used in a supervised setting, as we do, the GNN model can be used to learn task-specific messages to be transacted between nodes and/or edges in graphs.

We investigate the  usefulness of the proposed methods for segmenting tree-structures with an application to extract airway trees from CT data. {As both the methods are capable of obtaining any number of sub-graphs and not necessarily a tree, we make use of this feature to obtain a collection of sub-trees as predictions to the underlying trees without explicitly enforcing any tree constraints. This has the advantage of retrieving portions of airway trees which otherwise might be missed due to noise in the image data.}
We compare the MFN and GNN models to a baseline method that is similar to~\citet{lo2010vessel} that has been shown to perform well on a variety of CT datasets~\citet{pedersen2009danish, lo2012extraction, perez2016automatic} and {also to a 3D U-Net adapted to airway segmentation tasks~\citet{juarez2018automatic}.}


\section{Methods}\label{sec:meth}

In this section, we describe the task of graph refinement along with the underlying model assumptions. Based on this  model, we present two approaches to performing graph refinement using  MFNs and GNNs.

\subsection{Graph Refinement Model}
\label{sec:model}
Consider an over-complete, undirected, input graph, $\Gcal_\text{in}:\{\Vcal,\Ecal_\text{in}\}$, with nodes $i\in\Vcal : |\Vcal|=N$ associated with $F$-dimensional features, $\xbf_i \in \Rm^{F\times 1}$ collected into the node feature matrix, $\Xbf \in \Rm^{F \times N}$, and pairwise edges, $(i,j)\in\Ecal_\text{in}$, described by the input adjacency matrix, $\Abf_\text{in} \in \{ 0,1\} ^{N \times N}$. The goal of graph refinement is to recover a subgraph, $\Gcal$, with a subset of edges, $\Ecal \subset \Ecal_\text{in}$, described by the symmetric output adjacency matrix, $\Abf \in \{0,1\}^{N \times N}$. This subgraph corresponds to the structure of interest, like airway trees from chest data as studied in this work. We then seek a model, $f(\cdot)$, that can recover the subgraph from the input graph, $f: \Gcal_\text{in} \rightarrow \Gcal$.

\subsection{Mean-Field Networks}
\label{sec:mfn}
We next propose a probabilistic graph refinement model by introducing a random variable that captures the connectivity between any pair of nodes $i$ and $j$: $s_{ij} \in \{0,1\}$, with the probability of the corresponding connection given as $\alpha_{ij} \in [0,1]$. For each node $i$, the binary random variables associated with its incident connections are collected into a node connectivity variable $\sbf_i = \{ s_{ij} \} : j = 1 \dots N $. At the graph level, all node connectivity variables are collected into a global connectivity variable, $\Sbf = [ \sbf_1 \dots \sbf_N]$. 

The graph refinement model is described by the conditional distribution, $p(\Sbf|\Xbf,\Abf_\text{in})$, where the node features, $\Xbf$, and input adjacency, $\Abf_\text{in}$, are observed from the data. We use the notion of node potential, $\phi_{i} (\sbf_i)$, and pairwise edge potential, $\phi_{ij}(\sbf_i,\sbf_j)$, to express the joint distribution $p(\Sbf,\Xbf,\Abf_\text{in})$ and relate it to the conditional distribution as
	\begin{align}
			\ln p(\Sbf| \Xbf,\Abf_\text{in}) =   \ln \frac{p(\Sbf, \Xbf,\Abf_\text{in})}{p(\Xbf,\Abf_{\text{in}})} &  \propto  \ln p(\Sbf, \Xbf,\Abf_\text{in}) \nonumber \\	
		&	= \sum_{i \in \Vcal} \phi_{i} (\sbf_i) + \sum_{(i,j) \in \Ecal_\text{in}} \phi_{ij}(\sbf_i,\sbf_j) -\ln Z 
    \label{eq:cliq}
\end{align}
where $\ln Z$ is the normalisation constant. For ease of notation, explicit dependence on observed data in these potentials is not shown. It can be noticed this model bears similarities with the hidden Markov random field (MRF) models written in terms of Gibbs potentials~\cite{sandberg2004markov} that have been previously used for image segmentation, where the joint distribution is approximated with unary and pairwise energy functions~\citep{zhang2001segmentation,orlando2014learning}. To design suitable potentials for graph refinement we model terms that contribute positively when the nodes or the pairwise connections are likely to belong to the subgraph, and less positively or even negatively otherwise.  

First, we propose a node potential that captures the importance of a given node in the underlying subgraph, $\Gcal$.
For each node $i\in\Vcal$, it is given as 
\begin{equation}
	\phi_i(\sbf_i) = \sum_{v=0}^{D} \beta_v \Im \Big [ \sum_{j} s_{ij} = v\Big ] +  \abf^T \xbf_i\sum_{j} s_{ij},
	\label{eq:phiN}
\end{equation}
where $\sum_j s_{ij}$ is the degree of node $i$ and $\Im[\cdot]$ is the indicator function. The parameters $\beta_v \in \Rm,\text{ } \forall \text{ } v = [0,\dots,D]$, can be seen as prior weighting on the degree per node. We explicitly model and learn this term for up to 2 edges per node and assume a uniform prior for $D>2$. Nodes with $D=0$ correspond to nodes that do not belong to the subgraph, $D=1$ are root or terminal nodes and $D=2$ are the most common nodes in the subgraph which are connected to a parent node and a child node. For these cases we explicitly learn the parameter $\beta_v \text{ } \forall \text{ } v = [0,1,2]$. {The cases $D > 2$ are unlikely except in cases of bifurcations  ($D = 3$) and trifurcations ($D=4$) which are assumed to have a uniform prior.} Further, in the second term, a weighted combination of individual node features is computed using the parameter $\abf\in \Rm^{F \times 1} $ to represent the contribution of each feature to the node's importance. A node's importance, and hence its contribution to the node potential, is made dependent on its degree as seen in the second term in Equation~\eqref{eq:phiN}. That is, a node with more connections is more important to the subgraph and it contributes more to the node potential.

Secondly, we propose a pairwise potential that captures the relation between pairs of nodes and reflect their affinity to be connected in the underlying subgraph, $\Gcal$. For each pair of nodes $i,j \in \Vcal$, it is given as 
\begin{align}
	\phi_{ij}(\sbf_i,\sbf_j) =& \lambda \big( 1-2|s_{ij} - s_{ji}| \big ) \nonumber \\
	+& (2s_{ij}s_{ji}-1) \Big [ \boldsymbol{\eta}^T|\xbf_i-\xbf_j|_{e} + \boldsymbol{\nu}^T(\xbf_i \circ \xbf_j)\Big]. 
	\label{eq:phiE}
\end{align}
The first term in Equation~\eqref{eq:phiE} multiplied by $\lambda$ ensures symmetry in connections between nodes, i.e, for nodes $i,j$ it encourages $s_{ij} = s_{ji}$. As the distance between node features can be a useful indicator of existence of edge connections, a weighting of the absolute difference between nodes for each feature dimension, denoted as the element-wise norm $|\cdot|_{e}$, is computed using the parameter $\boldsymbol{\eta} \in \Rm^{F \times 1}$. 
The element-wise node feature product term $\boldsymbol{\nu}^{T}(\xbf_i \circ \xbf_j)$ computes a combination of the joint pairwise node features weighted by $\boldsymbol{\nu} \in \Rm^{F \times 1}$. The second term in Equation~\eqref{eq:phiE} is multiplied with $(2s_{ij}s_{ji}-1)$ to ensure that the contribution to the pairwise potential is positive when both nodes $i$ and $j$ are connected to each other, otherwise, the contribution is negative.

Returning to the posterior distribution, we note that except for in trivial cases, it is intractable to estimate $p(\Sbf|\Xbf,\Abf_\text{in})$ in Equation~\eqref{eq:cliq} and we must resort to approximating it. We take up the variational mean field approximation (MFA)~\citep{jaakkola1998improving}, which is a structured approach to approximating $p(\Sbf|\Xbf,\Abf_\text{in})$ with candidates from a class of simpler distributions$: q(\Sbf) \in \Qcal$. This approximation is performed by iteratively minimizing the exclusive Kullback-Leibler divergence~\citep{jaakkola1998improving}, or equivalently maximising the evidence lower bound (ELBO) or the variational free energy, given as
\begin{equation}
 \Fcal(q_\Sbf) = \ln Z+ \Em_{q_{\Sbf}} \Big [ \ln p(\Sbf| \Xbf,\Abf_\text{in}) - \ln q(\Sbf) \Big ],
 \label{eq:elbo}
\end{equation}
where $\Em_{q_{\Sbf}}$ is the expectation with respect to the distribution ${q_{\Sbf}}$. In MFA, the class of approximating distributions, $\Qcal$, are constrained such that $q(\Sbf)$ can be factored further. In our model, we assume that the existence of connection between any pair of nodes is independent of the other connections, which is enforced by the following factorisation: 
\begin{equation}
q(\Sbf) = \prod_\text{i=1}^N \prod_\text{j=1}^N q_{ij}(s_{ij}),  
	\label{eq:mfaFact}
\end{equation}
\begin{equation}
	\text{where, }
q_{ij}(s_{ij}) = \begin{cases}
\alpha_{ij} &\qquad \text{if }s_{ij} = 1 \\
(1-\alpha_{ij}) &\qquad \text{if }s_{ij} = 0
\end{cases},
\label{eq:mfa}
\end{equation}
with $\alpha_{ij}$ as the probability of connection between nodes $i$ and $j$.

Using the potentials from~\eqref{eq:phiN} and~\eqref{eq:phiE} in~\eqref{eq:elbo} and taking expectation with respect to $q_\Sbf$, we obtain the ELBO in terms of~$\alpha_{ij} \text{ }\forall\text{ } i,j=[1,\dots,N]$. By differentiating this ELBO with respect to any individual $\alpha_{kl}$, as elaborated in Appendix~\ref{sec:app}, we obtain the following update equation for performing MFA iterations. At iteration $(t+1)$, for each node $k$,
\begin{equation}
	\mathlarger \alpha_{kl}^{(t+1)} = \mathlarger \sigma({\gamma_{kl}}) = \frac{1}{1+\exp^{-\gamma_{kl}}} \text{ } l \in \Ncal_k,
    \label{eq:mfaUp}
\end{equation}
where $\mathlarger \sigma(\cdot)$ is the sigmoid function, $\Ncal_k$ are the $L$ neighbours of node $k$, and 
\begin{align}
	& \mathlarger \gamma_{kl} = 
\prod_{j \in \Ncal_k \setminus l} \big(1-\alpha_{kj}^{(t)}\big) \Big\{ \sum_{m \in \Ncal_k \setminus l} \frac{\alpha_{km}^{(t)}}{(1-\alpha_{km}^{(t)})} \Big[ (\beta_2-\beta_1) & \nonumber \\ 
	&	- \beta_2 \sum_{n \in \Ncal_k \setminus l,m} \frac{\alpha_{kn}^{(t)}}{(1-\alpha_{kn}^{(t)})}\Big]
   + \big(\beta_1-\beta_0 \big) \Big\} + \abf^T \xbf_k
     \nonumber \\ 
	& +	(4\alpha_{lk}^{(t)}-2)\lambda + 2\alpha_{lk}^{(t)}\big( \boldsymbol{\eta}^T|\xbf_k-\xbf_l|_e + \boldsymbol{\nu}^T(\xbf_k \circ \xbf_l) \big).
    \label{eq:mfa}
\end{align}
After each iteration $(t)$, the MFA procedure outputs predictions for the global connectivity variable, ${\alphabf}^{(t)}$, with entries $\mathlarger\alpha_{kl}^{(t)}$ given in Equation~\eqref{eq:mfaUp}. These MFA iterations are performed until convergence; a reasonable stopping criterion is when the increase in ELBO between successive iterations is below a small threshold. MFA guarantees convergence to a local optimum and the number of iterations $T$ can vary between tasks~\cite{wainwright2008graphical}. In such a case, $T$ can be treated as a hyperparameter that is to be tuned as described in Section~\ref{sec:mfnParam}. However, in most applications with MFNs $T$ is a small value ( $ < 50$) (Li et al. 2014).

It can be noticed that the MFA update procedure described in Equation~\eqref{eq:mfaUp} and Equation~\eqref{eq:mfa} resemble the computations in a feed-forward neural network. The predictions from iteration $(t)$, ${\alphabf}^{(t)}$, are combined and passed through a non-linear activation function, a sigmoid in our case, to obtain predictions at iteration $(t+1)$, ${\alphabf}^{(t+1)}$. This allows us to perform $T$ iterations of MFA with a $T$-layered network based on the underlying graphical model. This can be seen as the mean field network (MFN)~\citep{li2014mean}. 
The parameters of the MFN model, $[\lambda, \mathbf{\beta, a, \eta,\nu}]$, form weights of such a network and are shared across all layers. Given this setting, parameters for the MFN can be learned by back-propagating any suitable loss, $\Lcal({\alphabf}, \Abf_r)$, computed between the predicted global connectivity variable at the final iteration $\alphabf={\alphabf}^{(T)}$ and the reference adjacency, $\Abf_r$. 
We recover a symmetric adjacency matrix from the predicted global connectivity variable as $\Abf = \Im[({\alphabf} > 0.5) \land (\alphabf^T > 0.5)]$, because symmetry is not enforced on the predicted global connectivity variable i.e, the equality $\alpha_{ij} = \alpha_{ji}$ does not always hold. This is because of the MFA factorisation in Equation~\eqref{eq:mfaFact} where we assume connections between pairs of nodes to be independent of other connections.  Details of the MFN training procedure are presented in Section~\ref{sec:train}.

\subsection{Graph Neural Networks}
\label{sec:gnn}

With MFN in Section~\ref{sec:mfn}, we presented a hand-crafted model to perform graph refinement. In this section, we investigate if the messages transacted between nodes according to Equations~\eqref{eq:mfaUp} and~\eqref{eq:mfa} in the MFN can be learnt in a supervised setting using Graph Neural Networks.

In this work, we extend the preliminary work in~\citep{selvan2018extraction}, where graph refinement was performed using an encoder that learnt node embeddings, and a simple pairwise decoder that predicted edge probabilities from the node embeddings. We now propose the use of an edge-GNN based encoder and a single layer perceptron based decoder that predicts edge probabilities from the learnt edge embeddings. {This model has similarities with relational networks and interaction networks like in~\citep{battaglia2016interaction} mainly in the choice of using multi-layer perceptron (MLP) based node and edge aggregation strategies but differ in two essential aspects. Firstly, the interaction networks in~\citep{battaglia2016interaction} assume known graph structures whereas we assume an over-connected (in extreme cases, can consider fully connected graph) and model a graph refinement task to retrieve the underlying graph structure. Secondly, the decoder in the proposed GNN model predicts edge probabilities which in the interaction network setting would correspond to predicting the interaction between objects. This being said, both the models still are special cases of the more general GNNs proposed in~\citep{scarselli2009graph}.}

The graph refinement task, as formulated in Section~\ref{sec:model}, provides a conducive setting to use GNN based edge prediction, $f: \Gcal_\text{in} \rightarrow \Gcal$. 
{The GNN model, in our case, is used in a supervised setting to learn edge embeddings from which the subgraphs of interest can be reconstructed. Joint training of the encoder-decoder pair yields an encoder that first maps the input node features, $\Xbf$, to node embeddings, then computes the corresponding edges based on the input adjacency matrix, $\Abf_\text{in}$, and obtains expressive edge embeddings. The perceptron based decoder finally uses the learnt edge embeddings to predict the global connectivity variable, $\alphabf$.}

Following the notation in~\citep{kipf2018neural}, we present a GNN based encoder with a receptive field of two obtained using hyper-parameter tuning described in Sec.~\ref{sec:gnnParam}. The two GNN layers are identified by the superscripts:
\begin{align}
	\text{Node Embedding: }& & \hbf^{(1)}_j &&=&&g_n(\xbf_j) \label{eq:nodeEm} \\
	\text{Node-to-Edge mapping: }& &\hbf^{(1)}_{(i,j)} &&=&& g_{n2e}([\hbf_i^{(1)},\hbf_j^{(1)}]) \label{eq:n2e1} \\
	\text{Edge-to-Node mapping: }& &\hbf^{(2)}_j &&=&& g_{e2n}(\sum_{i}^{N_j}\hbf_{(i,j)}^{(1)}]) \label{eq:e2n1} \\ 
	\text{Node-to-Edge mapping: }& &\hbf^{(2)}_{(i,j)} &&=&& g_{n2e}([\hbf_i^{(2)},\hbf_j^{(2)}]) \label{eq:n2e2}
\end{align}
where each of the $g_{\dots}(\cdot)$ above is a 2-layered MLP with rectified linear unit activations, dropout~\citep{srivastava2014dropout} between the two hidden layers, skip connections~\citep{he2016deep} and layer normalisation~\citep{ba2016layer}. Equation~\eqref{eq:nodeEm} describes the node embedding corresponding to the first GNN, $\hbf_j^{(1)}$. The MLP, $g_n(\cdot)$, has $F$ input channels and $E$ output channels transforming the $F-$dimensional input node features into $E-$dimensional node embedding. The edge embedding, $\hbf_{(i,j)}^{(1)}$ for a pair of nodes, $(i,j)$ is obtained by simply concatenating the corresponding node features and propagating these features through the edge MLP, as described in Equation~\eqref{eq:n2e1}. The edge MLPs, $g_{n2e}(\cdot)$ have $2E$ input channels and $E$ output channels. Going from these edge embeddings to node representation is performed by simply summing over all the incoming edges to any given node $j$ from its neighbourhood, $\Ncal_j$ according to Equation~\eqref{eq:e2n1}. With this operation of summing over the neighbourhood of node $N_j$ is where the state of node $j$ is updated with its neighbourhood features (along edges $(i,j)$). Updated node embeddings are obtained by propagating these node features through the second node MLP, $g_{e2n}(\cdot)$ with $E$ input and output channels, as described in Equation~\eqref{eq:e2n1}. The second edge MLP, $g_{n2e}(\cdot)$ also has $2E$ input and $E$ output channels. Finally, the output from the encoder, $\hbf^{(2)}_{(i,j)}$, in Equation~\eqref{eq:n2e2} is the $E-$dimensional edge embedding which is used to predict the edge probabilities with a simple decoder. 

The perceptron based decoder is given as:
\begin{equation}
\mathlarger \alpha_{ij} = \mathlarger\sigma(g_{dec}(\hbf^{(2)}_{(i,j)}))
	\label{eq:dec}
\end{equation}
where $g_{dec}$  is a linear layer with bias and one output unit, and $\mathlarger \sigma(\cdot)$ is the sigmoid activation function. This decoding operation converts the $E-$dimensional edge embedding into a single scalar for each edge and the sigmoid function yields the corresponding edge probability, $\alpha_{ij}$. These,  $\alpha_{ij}$'s, form entries of the predicted global connectivity variable, $\alphabf$, similar to the predictions obtained from the MFN in Equation~\eqref{eq:mfaUp}. As with MFN, the GNN model loss is computed based on edge probability predictions, $\alphabf$, and the reference adjacency matrices, $\Lcal(\alphabf,\Abf_r)$.

Although the GNN model described above is for individual nodes and edges, these can be vectorised for faster implementation~\citep{kipf2018neural}. Also, the receptive field of the encoder can be easily increased by stacking more GNNs i.e, successive repetition of pairs of Equations~\eqref{eq:e2n1} and~\eqref{eq:n2e2}. 

\begin{figure*}[t]
        \centering{
                \includegraphics[width=0.79\textwidth,height=0.17\textheight]{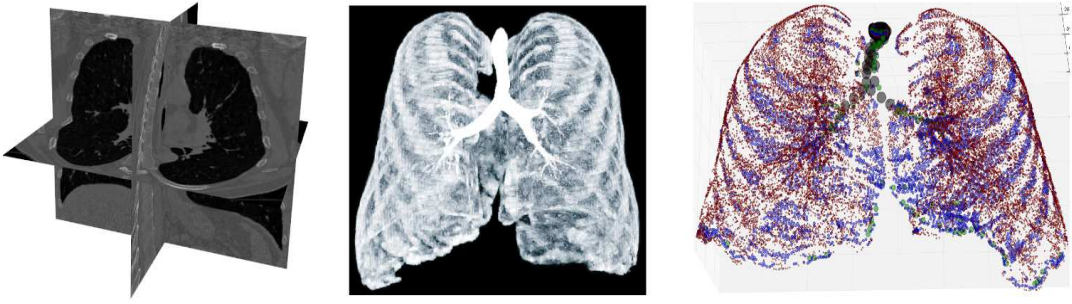}}
        \caption{The preprocessing to transform the input image (left) into a probability image (center) and then into graph format (right). Nodes in the graph are shown in scale (as different colours) to capture the variations in their local radius.}
    \label{fig:preproc}
\end{figure*}

\subsection{Loss Function}
\label{sec:train}

Both, MFN and GNN, models output predictions for the global connectivity variable, $\alphabf$, which has entries corresponding to the probability of pairwise connections. From a loss point of view, this is similar to a binary classification task, as the reference adjacency matrix, $\Abf_r$, has binary entries indicating the presence of edges in the underlying subgraph of interest. In most applications, the graphs are sparse as the edge class is in minority. To overcome challenges during training due to such class skew we use the Dice loss~\citep{milletari2016v} for optimising both the models, for its inherent ability to account for class imbalance. Dice loss is given as:
\begin{equation}
	\Lcal({\alphabf}, \Abf_r) = 1-\frac{2\sum_{i,j=1}^N \alpha_{ij}A_{ij}} {\sum_{i,j=1}^N {\alpha}_{ij}^2 + \sum_{i,j=1}^N  {A}_{ij}^2} ,
        \label{eq:dice}
\end{equation}
where $A_{ij}$ are the individual binary entries in the reference adjacency matrix.

\section{Experiments and Results}\label{sec:exp}
%

\subsection{Airway Tree Extraction as Graph Refinement}

Both the MFN and GNN models presented are general models that can be applied to broader graph refinement tasks with slight modifications. Here we present extraction of airway centerlines from volumetric CT images as a graph refinement task and describe the specific features used for this application. 

\subsubsection{Preprocessing}
\label{sec:preproc}
The image data is preprocessed to convert it into a graph format. First, the 3D CT image data is converted into a probability map using a trained voxel classifier according to~\citep{lo2010vessel}. This step converts intensity per voxel into a probability of that voxel belonging to the airway lumen. These probability images are transformed to a sparse representation using a simple multi-scale blob detector. Next, we perform Bayesian smoothing, with process and measurement models that model individual branches in an airway tree, using the method of~\citep{selvan2017extraction}. This three-step pre-processing procedure yields a graph output of the input image data, as illustrated in Figure~\ref{fig:preproc}. Each node in this graph is associated with a $7-$dimensional Gaussian density comprising of spatial location $\xbf_p=[x,y,z]$ in the image, local radius ($r$), and orientation $(v_x,v_y,v_z)$, such that $\xbf_i = [\xbf^i_{\mu}, \xbf^i_{\sigma^2}]$, comprising mean, $\xbf^i_{\mu}\in \Rm^{7\times 1}$, and variance for each feature, $\xbf^i_{\sigma^2}\in \Rm^{7\times 1}$. The node features are normalized to be in the range $[-1,1]$ for each scan to make all features of equal importance at input and to help in training the models.

\begin{figure*}[t]
\centering
	\includegraphics[width=0.85\textwidth]{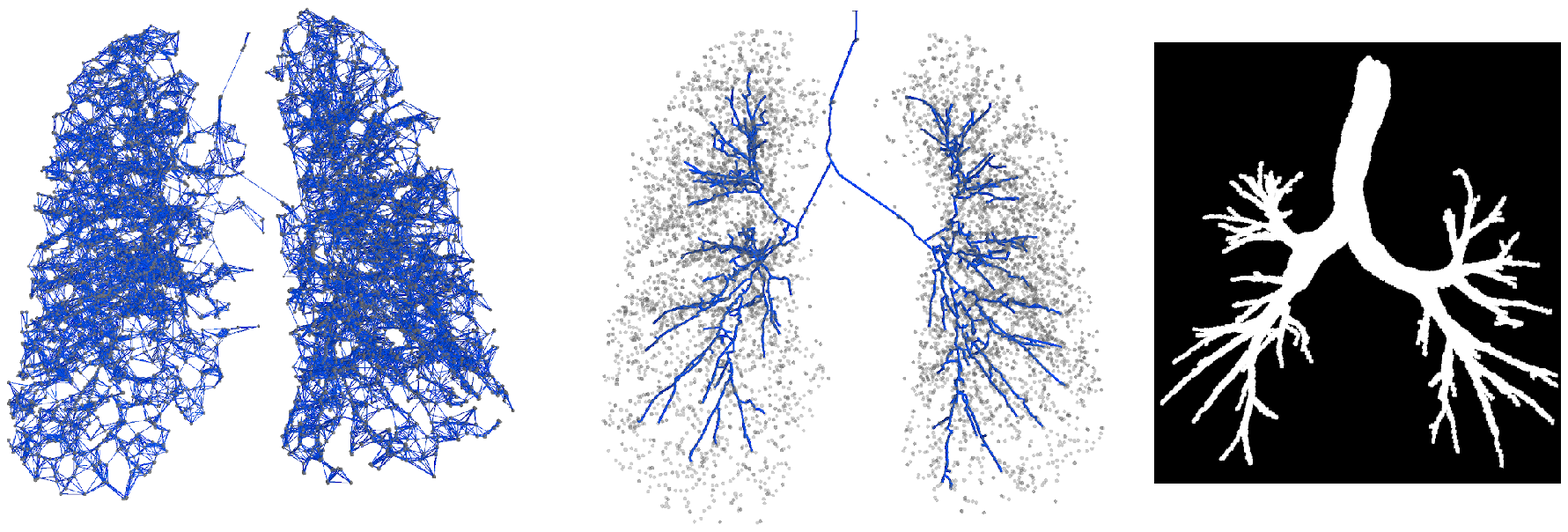}
	\caption{Input graph derived from a chest scan depicting the initial connectivity based on $\Abf_\text{in}$ between nodes (left). Nodes of the input graph (grey dots) overlaid with connections derived from the reference adjacency matrix, $\Abf_r$ (center). Binary volume segmentation obtained from the reference adjacency matrix and the corresponding node features (right). Note that due to visualisation artifacts introduced by viewing 3-d image projected into 2-d the nodes and the edges might not be very clear.}
	\label{fig:ref}
\end{figure*}

		The nodes in a graph that represent an airway tree are expected to have a certain behaviour. Nodes along a branch of airway will have a parent node and a sibling node. If the node is either a terminal node of an airway then it only has a parent node. In cases of bifurcations or trifurcations, the most neighbours a node can be connected to is three or four respectively. Taking this behaviour into account we allow for a larger number of possible neighbourhood of $10$. To obtain an initial connectivity, $\Abf_\text{in}$, we connect nodes to their 10 nearest neighbours based on spatial distance. These over-connected graphs, with node features $\xbf_i \in \Rm^{14 \times 1}$, are the input graphs for both models. Figure~\ref{fig:ref} (left) visualises an over-connected input graph.

\subsubsection{Adapting the MFN model}
\label{sec:adaptMFN}

\begin{figure}[h]
		\centering
		\includegraphics[width=0.25\textwidth]{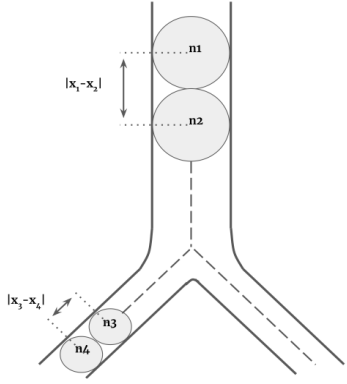}
		\caption{Schematic visualisation of the relative position features $|\xbf_i-\xbf_j|$ at two different scales inside an airway lumen. Nodes $n_1,n_2$ can be seen as nodes in a larger airway whereas the nodes $n_3,n_4$ depict nodes in smaller airways.}
		\label{fig:normRadius}
\end{figure}

The node and pairwise potentials in equations~\eqref{eq:phiN} and~\eqref{eq:phiE} are general and applicable to commonly encountered trees. Due to the nature of features extracted for the nodes in Section~\ref{sec:preproc}, one of the terms in the pairwise potential in Equation~\eqref{eq:phiE} requires a minor modification. The factor in Equation~\eqref{eq:phiE} associated with $\boldsymbol{\eta}$ is the element-wise absolute difference in node features, $|\xbf_i - \xbf_j|_e$. 
		The distance between two nodes inside a larger airway is larger than the distance between two nodes inside a smaller airway, as illustrated in Figure~\ref{fig:normRadius}. To make the relative position feature more meaningful we normalise the relative position with the average radius of the nodes, i.e.,  $|\xbf_p^i-\xbf_p^j|_e/(r^i+r^j)$, as the relative positions of each pair of connected nodes is proportional to their radii. The adaptions presented here incorporate specific domain knowledge pertaining to airways. In order to adapt this model to other tasks, for instance to extract airways, it will require additional considerations specific to that task. This can be a strength of the model by allowing inclusion of informative priors when available.

\subsubsection{Reference Adjacency Matrices}

Reference adjacency matrices are obtained from the reference segmentations using the preprocessing procedure described in Section~\ref{sec:preproc}. 
The extracted nodes and edges that are inside the corresponding reference segmentations are connected using a minimum spanning tree algorithm to obtain a single connected tree, yielding reference adjacency matrices that are used for training both the GNN and MFN models. A sample input graph along with the connections based on the reference adjacency matrix is shown in Figure~\ref{fig:ref} (center) .

\subsection{Data}

The experiments were performed on 3-D, low-dose CT, chest scans from the Danish lung cancer screening trial~\citep{pedersen2009danish}. All scans have voxel resolution of approximately $0.78\times 0.78 \times 1$ mm$^3$.  {We use two non-overlapping sets of $32$ scans and $100$ scans for evaluation and training purposes}. The $32$ scans in the first subset have reference segmentations that are treated as the ground truth for the purpose of evaluations, referred to as the \emph{reference dataset}. These reference segmentations are obtained by combining results from two previous airway segmentation methods~\citep{lo2010vessel,lo2009airway} that are corrected by an expert user. First of these methods uses a trained voxel classifier to distinguish airway regions from the background to yield probability images, and airway trees are extracted with region growing on these probabilities using an additional vessel similarity measure~\citep{lo2010vessel}. The second method extracts airways by extending locally optimal paths on the same probability images~\citep{lo2009airway}. %
{The second set comprising 100 scans has automatic segmentations obtained using ~\citep{lo2009airway}. As the reference dataset is relatively small, we use the second set of 100 scans to perform pre-training and to tune hyperparameters of both the models, referred to as the \emph{pre-training dataset}.} The specific choice of hyperparameter selection procedure for MFN model is presented in Section~\ref{sec:mfnParam} and for GNN model in Section~\ref{sec:gnnParam}.

\subsection{Evaluation}

\label{sec:bin}

The output of graph refinement models yields connectivity information about the airway centerlines. For evaluation purposes, we convert the predicted subgraph into a binary segmentation. This is done by drawing binary voxels within a tubular region that interpolates the radii of the nodes, along edges given by $\Abf = \Im[(\alphabf > 0.5) \land (\alphabf^T > 0.5)]$. One such binary segmentation is visualised in Figure~\ref{fig:ref} (right). 

Comparison of the graph refinement performance of the MFN and GNN models is done based on computing Dice similarity coefficient using the predicted and reference adjacency matrices
\begin{equation}
	Dice = \frac{2 | \Abf \circ \Abf_r | }{ |\Abf| + |\Abf_r| }.
	\label{eq:dice_acc}
\end{equation}

To evaluate the binary segmentations obtained using the procedure in Section~\ref{sec:bin}, centerline distance is used. Centerlines are extracted from the binary segmentations using a 3-D thinning algorithm~\citep{homann2007implementation} to be consistent in the evaluation of all comparing methods. The extracted centerlines are compared with the corresponding reference centerlines using an error measure that captures the average centerline distance. It is defined as:
\begin{align}
	d_{err} &= \frac{\sum_{i=1}^{N_{seg}} \min [d_E(c_i,C_{ref})]}{2N_{seg}} 
			+ \frac{\sum_{j=1}^{N_{ref}} \min [d_E(c_j, C_{seg})]}{2N_{ref}} \nonumber \\
			&=  \frac{d_{FP}+d_{FN}}{2}
			\label{eq:derr}
\end{align}

where the first factor $d_{FP}$ captures the errors due to possible false positive branches -- it is the average minimum Euclidean distance from segmented centerline points, $C_{seg}: |C_{seg}| = N_{seg}$, to reference centerline points, $C_{ref}: |C_{ref}| = N_{ref}$, --  and $d_{FN}$ captures the errors due to likely false negatives -- it is the average minimum Euclidean distance from reference centerline points to segmentation centerline points. 

We report two other commonly used measures in airway segmentation tasks, similar to those used in EXACT'09 challenge~\citep{lo2012extraction}. The fraction of tree length (TL) that is accurately detected, computed as
\begin{equation}
	\frac{L_{seg}}{L_{ref}}\times 100\%,
	\label{eq:treeLen}
\end{equation}
where $L_{seg}$ is the total length of accurately detected branches and $L_{ref}$ is the total length of all branches in the reference segmentation. Finally, the false positive rate (FPR) computed based on the number of centerline voxels outside the reference segmentation $N_w$ is given as,
\begin{equation}
	\frac{N_w}{N_{seg}}\times 100
	\label{eq:fpr}
\end{equation}
where $N_{seg}$ number of voxels in the output centerline. Note, however, that the EXACT evaluation uses binary segmentation and not the centerline to compute FPR.

{Evaluation of the graph refinement models and the baselines were performed using an $8-$fold cross validation procedure using the 32 scans in the \emph{reference} dataset, with $28$ scans for training and $4$ for testing within each fold.}

\begin{figure}[t]
\centering
\includegraphics[width=0.45\textwidth]{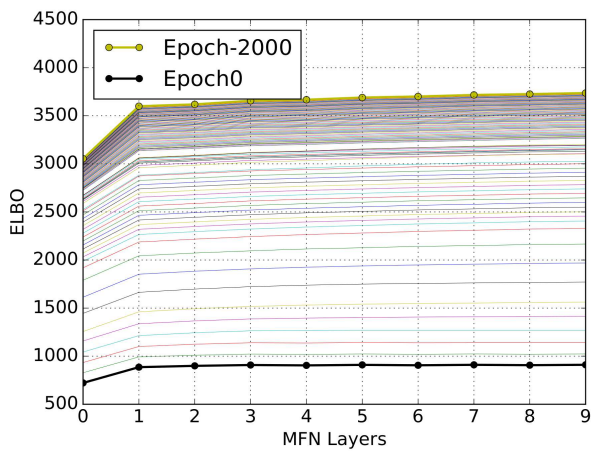}
\caption{The evolution of ELBO with each epoch and across each MFN layer for the MFN model. A clear trend of increase in ELBO within each epoch and across epochs is seen.}
\label{fig:elbo}
\end{figure}

\subsection{Training the models}

{Training of both MFN and GNN models was performed in three stages: hyperparameter tuning, pre-training and final model training, using the Dice loss in Equation~\eqref{eq:dice}. Hyperparameters such as the number of layers, training iterations and learning rate were tuned, and pre-training of both models was performed, using the \emph{pre-training} dataset. The model parameters were trained using the 32 scans in the \emph{reference} dataset in a cross validation set up.}
 All experiments were performed using a GPU with $12$GB memory and the code was implemented in PyTorch. The AMSGrad variant of Adam optimizer was used for optimization~\citep{reddi2018convergence} with an initial learning rate of $0.005$.

\subsubsection{MFN parameters}
\label{sec:mfnParam}
The most important hyperparameter in the MFN model is the number of layers $T$, equivalently the number of MFA iterations. Based on our initial experiments of observing the evolution of ELBO on the \emph{pre-training} dataset, (see Figure~\ref{fig:elbo}) we set the number of MF iterations or equivalently the number of layers in MFN to $T=10$, based on the discussions in Section~\ref{sec:mfn}. The number of training epochs was set to $2000$. On average each graph has around $8000$ nodes which are divided into sub-images comprising $500$ nodes, such that all the nodes in the input graph are taken into account to reduce memory utilisation. From an adjaceny matrix point of view we treat these sub-images as $500\times 500$ blocks in a block diagonal matrix and ensure all nodes are taken into account. Batch size of $12$ images (comprising all sub-images corresponding to an input graph) was used in the training procedure.

\subsubsection{GNN model parameters}
\label{sec:gnnParam}
{Based on the \emph{pre-training} dataset, we designed an architecture for the GNN model comprising an encoder with a receptive field of $2$ as described in Section~\ref{sec:gnn}, obtained from the range $[1,\dots,10]$. Validation accuracy and validation loss on the \emph{pre-training} dataset used to obtain the optimal number of GNN layers is depicted in Figure~\ref{fig:gnnLayers}. Each of the MLPs, $g_{\dots}(\cdot)$, used in the encoder in Equations~\eqref{eq:nodeEm}--\eqref{eq:n2e2} has two hidden layers chosen from the set $\{1,2,3,4\}$ and the number of channels per layer parameter $E=8$ chosen from $\{4,8,16,\dots,256\}$. A dropout rate of $0.5$ was used between each layer in the MLPs, chosen from the set $\{0, 0.1, \dots, 0.9\}$. The number of training epochs for the GNN model was set to $500$. Batch size of $12$ was used during training. Note that the GNN model can handle entire graphs utilising efficient sparse matrix operations and we do not require to subsample the graph as in the case of MFN model, as described in Section~\ref{sec:mfnParam}.

\begin{figure}[t]
\centering
\centering
\includegraphics[width=0.45\textwidth]{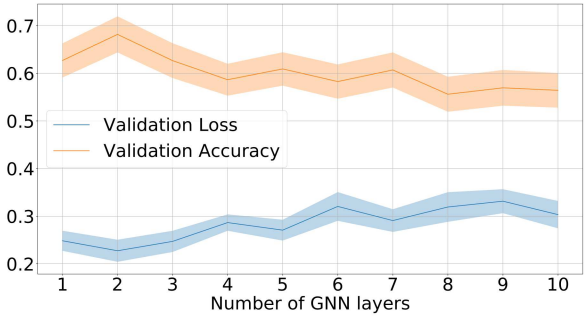}
\caption{{Influence of number of GNN layers on the validation accuracy and validation loss on the pre-training dataset. The best validation accuracy is obtained at L=2 which is the model used in this work.}}
\label{fig:gnnLayers}
\end{figure}

\subsection{Results}

\setlength{\tabcolsep}{3.5pt} 
\begin{table*}[t]

		\caption{Performance comparison of five methods: Region growing on probability images (Vox+RG), Bayesian smoothing merged with Vox+RG (BS+RG), UNet, MFN and GNN models. Dice similarity, centerline distances (d$_{FP}$, d$_{FN}$, d$_{err}$), fraction of tree length detected (TL) and false positive rate (FPR) are reported based on $8-$fold cross validation. Significant improvements when compared to other methods are shown in boldface. {Additionally, we also report the running time to train each of the models in a single fold. Note that the MFN and GNN models require additional preprocessing that is performed only once when preparing the graphs.}}
  \label{tab:res}

  \centering
	\scriptsize
	{
  \begin{tabular}{lccccccc}
    \toprule
	  {} & {Dice(\%)} & {$d_{FP}$(mm)} & {$d_{FN}$(mm)} & {$d_{err}$ (mm)} &  {TL(\%)}  & { FPR(\%)} & Time (m)\\
    \midrule
	  {Vox+RG} & -- &$2.937 \pm 1.005$ & $6.762 \pm 2.1042$ & $4.847 \pm  2.527$ & $73.2 \pm 9.9$ & $4.9 \pm 3.9$ & $90$\\
	  {BS+RG} & -- & $ 2.827 \pm 1.266$  & $4.601 \pm 2.002$ & $3.714 \pm 1.896$ & $73.6 \pm 6.1$ & $7.9 \pm 6.1$ & $105$\\
{	{UNet}} & -- & $ {3.540 \pm 1.316}$ & ${ 3.525 \pm 1.201}$ & ${3.532 \pm 1.259}$ & ${ 75.6 \pm 8.7}$ & ${6.5 \pm 3.3}$  & $5700$ \\
	  {MFN} & ${86.5 \pm 2.5}$ & $3.608 \pm 1.360$ & $ 3.116 \pm 0.632$ & $3.362 \pm 1.297$ & $74.5 \pm 6.7 $ & $ 8.6 \pm 5.4 $  & $60+35$\\
	  {GNN} & $84.8 \pm 3.3 $& $\mathbf{2.216 \pm 0.464}$ & $\mathbf{2.878 \pm 0.505}$ & $\mathbf{2.547 \pm 0.587}$ & $\mathbf{81.9 \pm 7.3}$ & $ {7.8 \pm 4.6}$ & $60+12$ \\
    \bottomrule
  \end{tabular}

}
\end{table*}

\label{sec:res}
\begin{figure}[t]
\centering
\includegraphics[width=0.5\textwidth]{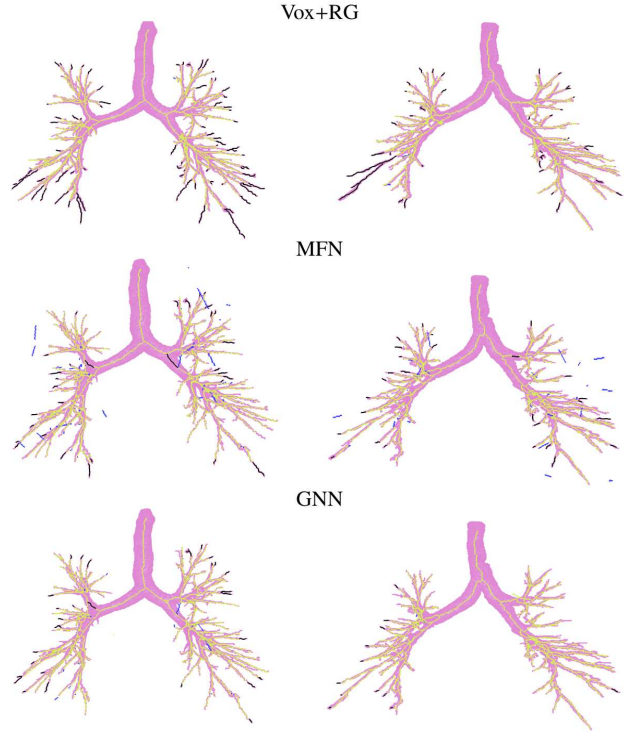}
	\caption{ Predicted centerlines for two test cases (along each column) from Vox+RG, MFN and GNN models overlaid with the reference segmentation (pink surface). In each case different colours are used to show true positive (yellow), false positive (blue) and false negative (black) branches.} 
\label{fig:res}
\end{figure}

We compare the performance of the MFN and GNN models to each other, with a baseline airway extraction method that uses region growing on probability images obtained using a voxel classifier, denoted \textbf{Vox+RG} and {3D UNet adapted for airway segmentation tasks~\citep{juarez2018automatic}, denoted \textbf{UNet}. The UNet model used for comparison closely follows the one reported in~\citep{juarez2018automatic} operating on five resolution levels, with elastic deformation based data-augmentation and optimising dice loss on the voxel predictions.} {The Vox+RG method is similar to the method in~\citep{lo2010vessel}, which was one of the top performing methods in EXACT'09 Challenge scoring the best FPR, had a reasonable tree completeness and was in the top five performing methods in TL measure}. Further, as the input to both graph refinement methods were nodes processed using the Bayesian smoothing method in~\citep{selvan2017extraction}, we also report the results for the Bayesian smoothing method. The output of Bayesian smoothing method is a collection of branches and not a complete segmentation; we merge its predictions with results of {Vox+RG} as in~\citep{selvan2017extraction}, denoted \textbf{BS+RG}.
\begin{figure}[h]
\centering
\includegraphics[width=0.69\linewidth]{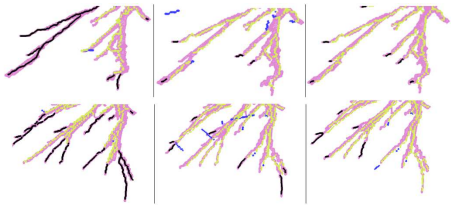}
\caption{{Comparison of the segmented branches in the lower right lobe for the Vox+RG, MFN and GNN models, respectively along the rows, for two cases. In each case different colours are used to show true positive (yellow), false positive (blue) and false negative (black) centerlines of branches; the reference segmentation is shown in pink.}}
\label{fig:zoom}
\end{figure}

Parameters of the region growing threshold of Vox+RG and BS+RG are tuned to optimise the average centerline distance in Equation~\eqref{eq:derr} using $8-$fold cross validation procedure on the \emph{reference} dataset. In Table~\ref{tab:res}, error measures for the proposed methods and the baselines are reported. Test set centerline predictions for two cases along with the reference segmentations for Vox+RG and the two graph refinement models are visualised in Figure~\ref{fig:res}. {To further highlight the improvements in detecting small, challenging branches, Figure~\ref{fig:zoom} shows the extracted branches for the two cases in the right lower lobe. Vox+RG method misses several branches entirely and large portions of the tips of these branches (black), which are largely extracted by the MFN and GNN methods. The MFN model, however, introduces additional false positive detections (blue). {We observe a similar behaviour in all the cases, which is overall captured as the increased d$_{FN}$ for Vox+RG and a significantly smaller false positive error d${_{FP}}$ for the GNN model.}} 

Based on the centerline distance measure reported in Table~\ref{tab:res} we see that both the MFN and GNN models show significant overall improvement captured in $d_{err}$ ($p < 0.001$) when compared to Vox+RG method. Both graph refinement methods specifically show large and significant improvement in $d_{FN}$ ($p< 0.001$), indicating their capability to detect more branches than Vox+RG, which is also evident in Figure~\ref{fig:res} (right). There is no improvement in $d_{FP}$, when compared to Vox+RG, for the MFN model, whereas for the GNN model there is a significant improvement ($p < 0.001$). Further, both graph refinement models show significant improvement ($p < 0.01$) when compared to Vox+RG in the fraction of tree length (TL) that is detected. 

To isolate the improvements due to preprocessing using Bayesian smoothing method in~\citep{selvan2017extraction} on the graph refinement models as described in Section~\ref{sec:preproc}, we report the centerline error for the predictions from BS+RG.
From the centerline distance measure entries in Table~\ref{tab:res}, we notice that both the graph refinement models show large and significant improvement ($p<0.001$) when compared to BS+RG method reported in the second row. A similar improvement is observed in TL for both the graph refinement models. 

{When compared to the 3D UNet model, the MFN model shows a significant improvement in: d$_{err}$ ($p < 0.001$) and has higher FPR ($p < 0.001$). The GNN model when compared to the 3D UNet also shows a significant improvement: d$_{err}$ ($p < 0.001$) \% TL $(p < 0.001)$ and no significant improvement in FPR ($p = 0.575$).} 

When comparing the performance between the MFN and GNN models in Table~\ref{tab:res}, we see a significant improvement using the GNN model in all centerline distance measures: $d_{FP}, d_{FN}, d_{err}$ ($p < 0.05$). Further, as the two graph refinement models predict the global connectivity variable, $\alphabf$, this performance is quantified by computing the Dice similarity coefficient, in Equation~\eqref{eq:dice_acc}, and reported in the second column in Table~\ref{tab:res}. We see that the MFN model obtains a higher score when compared to the GNN model indicating that the MFN model is better at predicting pairwise node connectivity. {All the reported significance values are based on two sided paired sample $t-$tests.}

{Training time for each of the models to process a single fold is reported in the last column of Table~\ref{tab:res}. The CNN based UNet model takes about $4$ days to process a single fold and whereas the other two baselines using region growing, Vox+RG and BS+RG models, can be trained in under $100$ min. The two graph refinement models use around $30$ min, however, they incur an additional one time preprocessing cost to prepare the graph structured data which is in the order of $60$ min.}



\section{Discussion and Conclusions}\label{sec:disc}

Detecting small branches and overcoming occlusions due to pathology and/or noise in data, during extraction of airways from CT data is challenging. By posing tree extraction as a graph refinement task we presented an exploratory approach that, to a large extent, overcomes these challenges. {Two models for graph refinement based on mean-field networks and graph neural networks were presented which allowed to extract any number of sub-graphs which we utilised to obtain collection of sub-trees as predictions to the underlying tree structures}. The proposed methods were evaluated on chest CT data and compared to a baseline method that is similar to~\citep{lo2010vessel} and a 3D UNet adapted for airway segmentation tasks~\citep{juarez2018automatic}. The method in~\citep{lo2010vessel} was one of the top performing methods in EXACT'09 airway extraction challenge~\citep{lo2012extraction} and forms a useful baseline for comparison. 

Some existing airway segmentation methods also have taken up an exploratory approach. The most recent and relevant work in this regard is~\citep{bauer2015graph}, where candidate airway branches are obtained using a tube detection filter and tree reconstruction is performed as a two-step graph-based optimisation. Candidate airway branches form nodes of this graph and plausible edges between these nodes are worked out in the first step of the optimisation. In the second step of the optimisation, sub-trees below a certain score are pruned away. In comparison to~\citep{bauer2015graph}, the proposed graph refinement setting operates on nodes that are local regions of interest and reconstructs branches and connections between branches simultaneously from these nodes. This graph refinement framework takes up a more global approach to tree reconstruction, as it does not rely on thresholding local sub-trees. 

{
The input to the two graph refinement models was based on a preprocessing step that used the Bayesian smoothing method in~\citep{selvan2017extraction} as described in Section~\ref{sec:preproc}. To isolate the improvements due to preprocessing and graph refinement models, we report the centerline error for the predictions from the Bayesian smoother as BS+RG in Table~\ref{tab:res}. 
From the $d_{err}$ entries in Table~\ref{tab:res} we notice that both the graph refinement models show large and significant improvement ($p<0.001$) when compared to BS+RG method reported in the second row. Further, a similar improvement is observed in the fraction of tree length for both the graph refinement models. Based on these observations we claim that the large portion of performance improvements are primarily due to the graph refinement procedures given the fixed preprocessed node features.}

\subsection{MFN model}
\label{sec:discMfn}

The main contribution within the presented MFN framework is the novel formulation of airway extraction within a graph refinement setting and the formulation of the node and pairwise potentials in~\eqref{eq:phiN} and~\eqref{eq:phiE}. By designing the potentials to reflect the nature of tasks we are interested in, the MFN model can be applied to diverse applications. For instance, it has been showed that information from pulmonary vessels can be used to improve airway segmentation in~\citep{lo2010vessel}. Modeling potential functions that take this information into account and encode the relation between vessel and airway branches could be done with MFN. Also, semantic segmentation tasks that predict voxel-level labels can also be modeled in the MFN setting, bearing similarities with the models used in~\citep{orlando2014learning}.

The MFN model can be seen as an intermediate between an entirely model-based solution and an end-to-end learning approach. It can be interpreted as a structured neural network where the interactions between layers are based on the underlying graphical model, while the parameters of the model are learnt from data. This, we believe, presents an interesting link between probabilistic graphical models and neural network-based learning. 


\subsection{GNN model}
\label{sec:discGNN}

In~\citep{selvan2018extraction}, we introduced the GNN model for graph refinement tasks. In that work, however, the GNN model was used to learn node embeddings using node GNNs. A pairwise decoder was then used to predict edge probabilities from the learnt node embeddings. With our experiments we found the model to be performing inadequately. 
With the model presented here, in Section~\ref{sec:gnn}, we introduced edge GNNs in the encoder to explicitly represent the edges, in order to learn edge embeddings. By jointly training the encoder-decoder pair now, we use the learnt edge embeddings to predict the probability of edges, showing clear improvements compared to the node GNN model in~\citep{selvan2018extraction}. 

The graph encoder used in this work consists of two GNN layers, meaning that nodes of the GNN have access to messages from first and second order neighbourhoods. This receptive field can be further increased by adding GNN layers. A sufficiently deep GNN-based encoder should allow each node to receive messages from all other nodes with increasing computational expense. For the graph refinement task considered here, we observed a receptive field of two to be sufficient. The choice of this receptive field was based on initial experiments on the \emph{pre-training} dataset with our observations reported in Figure~\ref{fig:gnnLayers}. This variation in validation performance is consistent with previously reported influence of increasing number of GNN layers in ~\citep{kipf2016semi}, which can be attributed to the increase in the range of neighbourhood for nodes and the ensuing difficulty in training the model due to the increase in number of parameters, which could be alleviated by exploring recurrent architectures or models with skip-connections and/or gating}. 


\subsection{Comparison between MFN and GNN models}
\label{sec:conn}

The MFN model update Equations~\eqref{eq:mfaUp} and~\eqref{eq:mfa} reveal the message passing nature of the underlying inference procedure~\citep{wainwright2008graphical}. The state of each node i.e., the edge update message from node $k$ to node $l$ is dependent on their corresponding data terms and all neighbours of node $k$ except node $l$. These messages transacted in a $T-$layered MFN are hand-crafted based on the model in Equations~\eqref{eq:phiN} and~\eqref{eq:phiE} and deriving an analytical solution that guarantees an increase in ELBO. However, deriving such analytical solutions might not be feasible for all scenarios. 

As GNNs can be seen as generalisation of message passing based inference methods~\citep{gilmer2017neural,yoon2018inference}, with a capability of learning complex task-specific messages, an interesting connection with the MFN model can be made. Given sufficient training data, in principle, the GNN model should be able to learn  messages to approximate the same posterior density as the MFN model. This connection is confirmed based on the centerline error measures reported in Table~\ref{tab:res}, wherein we see the two graph refinement models perform at least equally well for the same task. 


The mean-field factorisation, according to Equation~\eqref{eq:mfaFact} that resulted in the MFN model, means the connections between nodes are independent of each other, which is a strong assumption resulting in asymmetric predicted adjacency matrices.  
And, as the GNN model is trained in a supervised setting using symmetric adjacency matrices, the model predicts symmetric adjacency matrices in most cases.

The GNN model is able to detect more missing branches than the MFN model as seen in Table~\ref{tab:res}. There is a reduction in $d_{FP}$ for the GNN model; this is due to several spurious and disconnected branches predicted by the MFN model. The GNN model predicts fewer disconnected edges, indicating that, perhaps, the model is able to learn that stand-alone, disconnected edges are unlikely in an airway tree. This is clearly captured in the visualisations in Figure~\ref{fig:res}. 

From a graph refinement perspective, we see the MFN model scores higher in dice similarity (second column of Table~\ref{tab:res}). This is contrary to the centerline distance performance but can be explained by noticing that each edge in the dice accuracy in Equation~\eqref{eq:dice_acc} has the same importance. That is, edges between nodes in branches of large and small radii have the same importance. However, a missing edge in a branch of large radius can contribute more to the centerline distance than a missing edge in a branch of smaller radius. 

The GNN model used here is more complex, with $3150$ tunable weights, than the MFN model, which has a small set of tunable parameters $[\lambda, \boldsymbol{a, \beta,\eta,\nu}]$ and in all 46 tunable weights. Each training epoch containing  $28$ training images for the MFN model takes about $2s$ and $1s$ for the GNN model. The implementation of the GNN model takes advantage of sparse matrix operations, for $O(|\Ecal_\text{in}|)$ computational complexity. A similar sparse implementation can further reduce the computation time for the MFN model.

\subsection{Limitations and Future Work}

The pre-processing performed in Section~\ref{sec:preproc} is one possible way of obtaining graphs from image data as demonstrated in this work. A natural next step is to use more powerful local feature extractors based on CNNs and  learn the initial graph extraction. Initial work involving sequential training of feature extraction using CNNs and GNNs for learning global connectivity has been proposed in~\citep{shin2019deep} for 2-D vessel segmentation tasks. A joint end-to-end training procedure that dynamically extracts graphs from image data and performs graph refinement is challenging, but an interesting direction. Such models, where CNNs would be used as local feature extractors and GNNs operating on sparse graphs to model the global connectivity could be useful also to reduce the massive memory footprints of CNN models in 3D volumes. 

                

In the MFN model, we currently only use a linear data term in the node potential, $\abf^T\xbf_i$ in~\eqref{eq:phiN}, and a pairwise potential, $\boldsymbol{\nu}^T(\xbf_i \circ \xbf_j)$ in~\eqref{eq:phiE}.  There are possibilities of using more complex data terms to learn more expressive features. 
{Additional potential terms can be envisioned which can model bifurcations or penalise stand-alone branches. For the latter case, a potential term such as:
\begin{equation*}
\phi_{ij}  = -\Gamma \times \mathbb{I}[D(i) == D(j) == 1]
\end{equation*}
				imposes a negative penalty, $\Gamma$, to the mean-field optimisation  when the degree, $D(\cdot)$, of the nodes $i$ and $j$ are equal to $1$. When two nodes $(i,j)$ have a single edge connecting them, it translates to a stand-alone edge. Incorporating a potential of this kind penalises stand-alone edges between nodes with a negative cost and could possibly remove spurious singe branches as the ELBO optimisation progresses.}


While the output of the GNN has fewer disconnected branches when compared to MFN predictions, the output in all cases is not a fully connected tree. Incorporating tree enforcing constraints, either in the loss function or, in the GNN model could be beneficial. {For instance, an additional loss component that enforces tree behaviour, can be introduced, 
\begin{equation}
\Lcal_t = \frac{(N_c-1)}{Nc}
\end{equation}
where $N_c$ is the total number of connected components in the predicted adjacency matrix. Note that $\Lcal_t = 0 $ when $N_c = 1$, $\Lcal_t=0.5$ when $N_c = 2$ and $\Lcal_t \approx 1$ when $N_c >> 1$. Thus, the joint loss becomes: 
\begin{equation}
\Lcal = \Lcal_{dice} + \epsilon \times \Lcal_{t}
\end{equation}
where the dice loss in Eq. (14) is indicated as $\Lcal_{dice}$ and an annealing factor $\epsilon \approx 0$ at the start of the optimisation and increases gradually can be used to introduce the tree constraint. Use of such an annealing scheme could allow more disconnected components at the start of the optimisation leading into larger, unified structures. One could use REINFORCE-type gradient updates for learning as $\Lcal_t$ is not differentiable~\citep{williams1992simple}}.

\subsection{Conclusion}

In this work, we presented exploratory methods for the extraction of tree-structures from volumetric data, with a focus on airway extraction, formulated as graph refinement tasks. We proposed two novel methods to perform graph refinement based on MFNs and GNNs. 

We evaluated the two methods in their ability to extract airway trees from CT data and compared them to two relevant baseline methods. With our experiments, we have shown that both the MFN and GNN models perform significantly better than the baseline methods on the average centerline distance measure. Between the MFN and GNN models, the GNN model is able to detect more branches with fewer false positives as shown with the fraction of tree length and false positive rate measures. We have also presented connections between the MFN and GNN models. GNNs are more complex models which can be seen as generalisation of MFN models, while the MFN models are simpler and can be viewed as structured GNNs based on underlying graphical models.  

\subsubsection*{Acknowledgements}
This work was funded by the Independent Research Fund Denmark (DFF), Netherlands Organisation for Scientific Research (NWO) and SAP SE.

\section*{Acknowledgements}

Funding: This work was funded by the Independent Research Fund Denmark (DFF); SAP SE, Berlin; and Netherlands Organisation for Scientific Research (NWO).

\appendix
\appendix
\section{}
\label{sec:app}
We detail the procedure for obtaining the mean field approximation update equations in~(6) and~(7) starting from the variational free energy in equation~(4). We start by repeating the expression for the node and pairwise potentials.

\subsubsection*{Node potential}
\begin{equation}
	\phi_i(\sbf_i) = \sum_{v=0}^{2} \beta_v \Im \Big [ \sum_{j} s_{ij} = v\Big ] +  \abf^T \xbf_i\sum_{j} s_{ij},
	\label{eq:phiN_a}
\end{equation}
\subsubsection*{Pairwise potential}
\begin{align}
	\phi_{ij}(\sbf_i,\sbf_j) &= (2s_{ij}s_{ji}-1) \Big [ \boldsymbol{\eta}^T|\xbf_i-\xbf_j|_e + \boldsymbol{\nu}^T(\xbf_i \circ \xbf_j)\Big] \nonumber \\
&+ \lambda \big( 1-2|s_{ij} - s_{ji}| \big ) 
	\label{eq:phiE_a}
\end{align}

The variational free energy is given as,
\begin{equation}
 \Fcal(q(\Sbf)) = \ln Z+ \Em_{q(\Sbf)} \Big [ \ln p(\Sbf| \Xbf, \Abf_\text{in}) - \ln q(\Sbf) \Big ].
 \label{eq:elbo_a}
\end{equation}
Plugging in~\eqref{eq:phiN_a} and~\eqref{eq:phiE_a} in~\eqref{eq:elbo_a}, we obtain the following:
\begin{align}
& \Fcal(q(\Sbf)) = \ln Z+ \Em_{q(\Sbf)} \Big [ \sum_{i \in \Vcal} \Big\{ \beta_0 \Im \big [ \sum_{j} s_{ij} = 0\big ] \nonumber \\
&+ \beta_1 \Im \big [ \sum_{j} s_{ij} = 1\big ]  + \beta_2 \Im \big [ \sum_{j} s_{ij} = 2\big ]+   \abf^T \xbf_i\sum_{j} s_{ij} \Big\} 
\nonumber \\
	&+ \sum_{(i,j) \in \Ecal_\text{in}} \Big\{\lambda \big( 1-2|s_{ij} - s_{ji}| \big ) + (2s_{ij}s_{ji}-1) \Big [ \boldsymbol{\eta}^T|\xbf_i-\xbf_j|_e 
\nonumber \\
&+ \boldsymbol{\nu}^T(\xbf_i \circ \xbf_j)\Big]\Big\} - \ln q(\Sbf) \Big ].
\end{align}
We next take expectation $\Em_{q(\Sbf)}$ using the mean-field factorisation that $q(\Sbf) = \prod_{i=1}^N \prod_{j\in \Ncal_i} q_{ij}(s_{ij})$ and the fact that $\Pr\{s_{ij}=1\} = \alpha_{ij}$ we simplify each of the factors :
\begin{align}
& \Em_{q(\Sbf)} \Big [  \beta_0 \Im \big [ \sum_{j} s_{ij} = 0\big ] \Big] \nonumber \\ 
&= \Em_{q_{{i1}}\dots q_{{iN}}} \beta_0\Im \big [ \sum_{j} s_{ij} = 0\big ] \Big] = \beta_0 \prod_{j\in \Ncal_i}  (1-\alpha_{ij}).
\end{align}
Similarly,
\begin{align}
\Em_{q(\Sbf)} \Big [  \beta_1 \Im \big [ \sum_{j} s_{ij} = 1\big ] \Big] =  \beta_1 \prod_{j\in \Ncal_i}  (1-\alpha_{ij}) \sum_{j\in \Ncal_i} \frac{\alpha_{im}}{ (1-\alpha_{im})}
\end{align}
and
\begin{align}
& \Em_{q(\Sbf)} \Big [  \beta_2 \Im \big [ \sum_{j} s_{ij} = 2\big ] \Big] \nonumber \\ & =  \beta_2 \prod_{j\in \Ncal_i}  (1-\alpha_{ij}) \sum_{m\in \Ncal_i} \sum_{n \in \Ncal_i\backslash m} \frac{\alpha_{im}}{(1-\alpha_{im})}\frac{\alpha_{in}}{(1-\alpha_{in})}.
\end{align}
Next, we focus on the pairwise symmetry term:
\begin{equation}
\Em_{q(\Sbf)} \Big [ \lambda \big( 1-2|s_{ij} - s_{ji}| \big ) \Big] =\lambda \big( 1-2(\alpha_{ij} +\alpha_{ji}) +4\alpha_{ij}\alpha_{ji}\big)
\end{equation}
Using these simplified terms, and taking the expectation over the remaining terms, we obtain the ELBO as,
\begin{align}	
	& \Fcal({q(\Sbf)}) =  \ln Z+ \sum_{i \in \Vcal} \prod_{j \in \Ncal_i} (1-\alpha_{ij}) \Big \{ \beta_0  
+ \sum_{m \in \Ncal_i} \frac{\alpha_{im}} {(1-\alpha_{im})} \Big[ \beta_1   \nonumber \\ 
& + \beta_2 \sum_{n \in \Ncal_i \setminus m} \frac{ \alpha_{in}}{(1-\alpha_{in})} \Big] +
\abf^T \xbf_i\sum_{j} \alpha_{ij} \Big\}
 + \sum_{i \in \Vcal} \sum_{j \in \Ncal_i} \Big\{ 4\alpha_{ij}\alpha_{ji} \nonumber \\
& + \lambda \big( 1-2(\alpha_{ij} +\alpha_{ji}) \big)  -\Big( \alpha_{ij} \ln {\alpha_{ij}}  + (1-\alpha_{ij}) \ln (1-{\alpha_{ij}}) \Big) 
 \nonumber \\
&
+ (2\alpha_{ij}\alpha_{ji}-1) \Big [ \boldsymbol{\eta}^T|\xbf_i-\xbf_j|_e + \boldsymbol{\nu}^T(\xbf_i \circ \xbf_j)\Big]\Big \}.
	\label{eq:elbo_aw_a}
\end{align}
We next differentiate ELBO in~\eqref{eq:elbo_aw_a} wrt $\alpha_{kl}$ and set it to zero. 
\begin{align}
	&	\frac{\partial\Fcal{(q(\Sbf))}}{\partial \alpha_{kl}} =  \prod_{j \in \Ncal_k \setminus l} \big(1-\alpha_{kj}\big) \Big\{ \sum_{m \in \Ncal_k \setminus l} \frac{\alpha_{km}}{(1-\alpha_{km})} \Big[ (\beta_2-\beta_1) 
	\nonumber \\
&- \beta_2 \sum_{n \in \Ncal_k \setminus l,m} \frac{\alpha_{kn}}{(1-\alpha_{kn})}\Big] + \big(\beta_1-\beta_0 \big) \Big\} 
    + (4\alpha_{lk}-2)\lambda \nonumber \\ 
    & + \abf^T \xbf_k + 2\alpha_{lk}\big( \boldsymbol{\eta}^T|\xbf_k-\xbf_l|_e + \boldsymbol{\nu}^T(\xbf_k \circ \xbf_l) \big) -\Big [\ln \frac{\alpha_{kl}}{1-\alpha_{kl}} \Big ] \nonumber \\
	&= 0
\end{align}
From this we obtain the MFA update equation for iteration $(t+1)$ based on the states from $(t)$,
\begin{equation}
	\mathlarger \alpha_{kl}^{(t+1)} = \mathlarger \sigma({\gamma_{kl}}) = \frac{1}{1+\exp^{-\gamma_{kl}}} \text{ } \forall \text{ } {k} = \{1\dots N\},\text{ } l \in \Ncal_k
    \label{eq:mfaUp_a}
\end{equation}
where $\mathlarger \sigma(.)$ is the sigmoid activation function, $\Ncal_k$ are the $L$ nearest neighbours of node $k$ based of positional Euclidean distance, and 
\begin{align}
	&\mathlarger \gamma_{kl} = 
\prod_{j \in \Ncal_k \setminus l} \big(1-\alpha_{kj}^{(t)}\big) \Big\{ \sum_{m \in \Ncal_k \setminus l} \frac{\alpha_{km}^{(t)}}{(1-\alpha_{km}^{(t)})}
	\Big[ (\beta_2-\beta_1) 
\nonumber \\
&- \beta_2 \sum_{n \in \Ncal_k \setminus l,m} \frac{\alpha_{kn}^{(t)}}{(1-\alpha_{kn}^{(t)})}\Big] + \big(\beta_1-\beta_0 \big) \Big\} 
    + \abf^T \xbf_k 
     \nonumber \\ 
    &
+ (4\alpha_{lk}^{(t)}-2)\lambda + 2\alpha_{lk}^{(t)}\big( \boldsymbol{\eta}^T|\xbf_k-\xbf_l|_e + \boldsymbol{\nu}^T(\xbf_k \circ \xbf_l) \big).
    \label{eq:mfa_a}
\end{align}

\bibliographystyle{model2-names.bst}\biboptions{authoryear}
\bibliography{wrapper}

\end{document}